\documentclass[a4paper, 10pt, twocolumn, twoside]{article}

\usepackage{ISARC}

\usepackage{lscape}
\usepackage{hologo}

\begin{document}

\linespread{0.5}

\title{Interoceptive Robots for Convergent Shared Control in Collaborative Construction Work}

\author{Xiaoshan Zhou$^{1}$, Carol C. Menassa$^{2}$, and Vineet R. Kamat$^3$}

\affiliation{
$^1$Ph.D. Candidate, Dept. of Civil and Environmental Engineering, Univ. of Michigan, Ann Arbor, MI 48109, USA.\\
$^2$Professor, Dept. of Civil and Environmental Engineering, Univ. of Michigan, Ann Arbor, MI 48109, USA.\\
$^3$Professor, Dept. of Civil and Environmental Engineering, Univ. of Michigan, Ann Arbor, MI 48109, USA.\\
}

\email{
\href{mailto:e.author1@aa.bb.edu}{xszhou@umich.edu}, 
\href{mailto:e.author1@aa.bb.edu}{menassa@umich.edu}
\href{mailto:e.author1@aa.bb.edu}{vkamat@umich.edu}
}

\maketitle 
\thispagestyle{fancy} 
\pagestyle{fancy}

\begin{abstract}
Building autonomous mobile robots (AMRs) with optimized efficiency and adaptive capabilities—able to respond to changing task demands and dynamic environments—is a strongly desired goal for advancing construction robotics. Such robots can play a critical role in enabling automation, reducing operational carbon footprints, and supporting modular construction processes. Inspired by the adaptive autonomy of living organisms, we introduce interoception, which centers on the robot’s internal state representation, as a foundation for developing self-reflection and conscious learning to enable continual learning and adaptability in robotic agents. In this paper, we factorize internal state variables and mathematical properties as “cognitive dissonance” in shared control paradigms, where human interventions occasionally occur. We offer a new perspective on how interoception can help build adaptive motion planning in AMRs by integrating the legacy of heuristic costs from grid/graph-based algorithms with recent advances in neuroscience and reinforcement learning. Declarative and procedural knowledge extracted from human semantic inputs is encoded into a hypergraph model that overlaps with the spatial configuration of onsite layout for path planning. In addition, we design a velocity-replay module using an encoder-decoder architecture with few-shot learning to enable robots to replicate velocity profiles in contextualized scenarios for multi-robot synchronization and handover collaboration. These “cached” knowledge representations are demonstrated in simulated environments for multi-robot motion planning and stacking tasks. The insights from this study pave the way toward artificial general intelligence in AMRs, fostering their progression from complexity to competence in construction automation.
\end{abstract}


\section{Introduction}
\label{sec:Introduction}

Automation in construction can significantly contribute to achieving sustainability goals \cite{zheng2023intelligent}. In particular, modular construction using offsite prefabrication and onsite assembly have the potential to address climate action by reducing operational carbon footprints compared to traditional onsite methods such as concrete mixing and curing \cite{yevu2023digital}. In additional to potential environmental benefits, automation can improve worker safety by mitigating exposure to hazardous conditions. For example, automated robotic machinery used for onsite tasks, such as assembly, reduces risks like falls from scaffolding, inhalation of welding fumes, and electrocution \cite{liu2024automatic}. Mobile robots, including forklifts, cranes, and dozers, are increasingly used to streamline offsite and onsite construction processes, enhancing efficiency and reducing reliance on human workers and operators in dangerous scenarios \cite{you2021earthwork}. However, the integration of robotics in construction introduces new challenges. Without effective coordination of multi-robot path planning and precise velocity control, risks such as struck-by hazards increase \cite{kim2020proximity}. Proper coordination is essential to prevent collisions while accommodating the dynamic nature of construction workflows and managing sequential task dependencies.

Mobile robots introduce new dynamics to construction work by reshaping the interactions between humans and machines. These robots can operate with programmed autonomy, performing tasks such as delivering materials across a precast factory or a site to support fabrication or assembly processes \cite{liu2024automatic}. Although they remain under the supervision of human workers, their autonomy shifts the relationship from subservience—where robots follow human instructions without intrinsic understanding—to augmentation, where robots act more like symbiotic co-workers with specialized capabilities.

This shift, however, can provoke apprehension among industry practitioners, often fueled by fear of the unknown \cite{liang2024ethics}. Concerns about the emergence of superintelligent systems, often referred to as singularity, suggest a future where robots could become uncontrollable and irreversible in their autonomy. Such fears are compounded by the opaque reasoning processes of robots and the asymmetry in knowledge between humans and machines, leading to uncertainty about whether these systems are truly beneficial and trustworthy. To address these challenges, this research advocates for the development of advanced cognitive architectures for mobile robots that prioritize human oversight and meaningful interaction. By incorporating human input and feedback, the goal is to enhance transparency, trust, and the reasoning capabilities of robots, particularly in areas of adaptability and self-reflection.

Many principles of robotics are inspired by how humans perceive, reason, and act in the physical world. For instance, robotic sensing mimics human sensory organs: cameras (RGB, depth, or infrared) parallel human vision, force sensors replicate touch, microphones simulate hearing, and joint encoders mirror proprioception. Furthermore, both humans and robots share functional constraints, such as perceptual receptive fields and energy dependence. However, a key distinction between humans and robots lies in adaptability and learning. Human workers are inherently adaptive, capable of consciously learning and adjusting to external changes. In contrast, robots are mechanical and purpose-built, often limited in their ability to make context-aware decisions or learn from others performing similar tasks. Additionally, robotic designs typically focus on replicating exteroception (sensing the external environment), while humans possess interoception—the ability to perceive internal states, such as emotion and allostasis. Interoception guides human decisions, helping to counterbalance disruptions, form extrapolated schemas from observed patterns, and prioritize actions based on a hierarchy of needs \cite{Pezzulo_2015}. Inspired by this, this research emphasizes the potential benefits of equipping mobile robots with interoceptive-like capabilities.

\begin{figure}[!htb]
    \centering
    \includegraphics[width=0.48\textwidth]{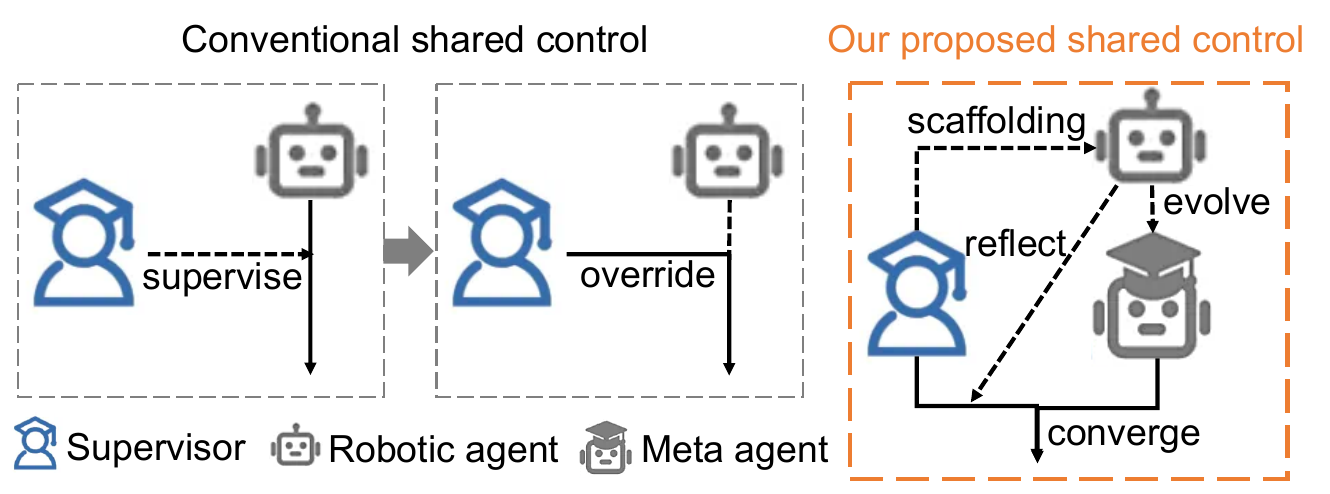}
    \caption{Shared control architectures}
    \label{fig_1}
\end{figure}

Interoception in mobile robots is a novel concept, and its conceptualization and implementation require a formal definition of the phenotypic characteristics—particularly the internal states—of mobile robots when supervised by humans or operating under human feedback in a closed-loop system. By “interoception”, we do not imply that robots experience genuine emotions, pain, or self-consciousness. Instead, this research defines cognitive dissonance as an abstracted form of a robotic internal state variable and factorizes it based on the intensity of human inputs to refine robotic autonomy. This is proposed with a new shared control architecture as illustrated in Figure ~\ref{fig_1}: Rather than allowing human interventions to override robotic decisions, the proposed shared control integrates human inputs as tributaries that converge with autonomous decisions to produce a collaborative output forwarded to the robot’s actuators. This architecture introduces a potential conflicting state between the robot’s internal beliefs and its executed actions. Such conflicts internally drive the robotic agent to reflect on its decisions, seeking insights from human feedback to identify potential biases or missed factors causing misalignment with human expectations. Through this reflective process, the robot engages in conscious-like learning, evolving into a meta-level understanding that reconciles discrepancies with its human counterparts. This framework is particularly valuable in spatio-temporal-sensitive situations in multi-robot systems navigating shared spaces and adapting to dynamic task sequences (e.g., delivering panels before façade installation) with accurate reproduction of planned velocity profiles to minimize downtime.

\begin{figure}[!htb]
    \centering
    \includegraphics[width=0.48\textwidth]{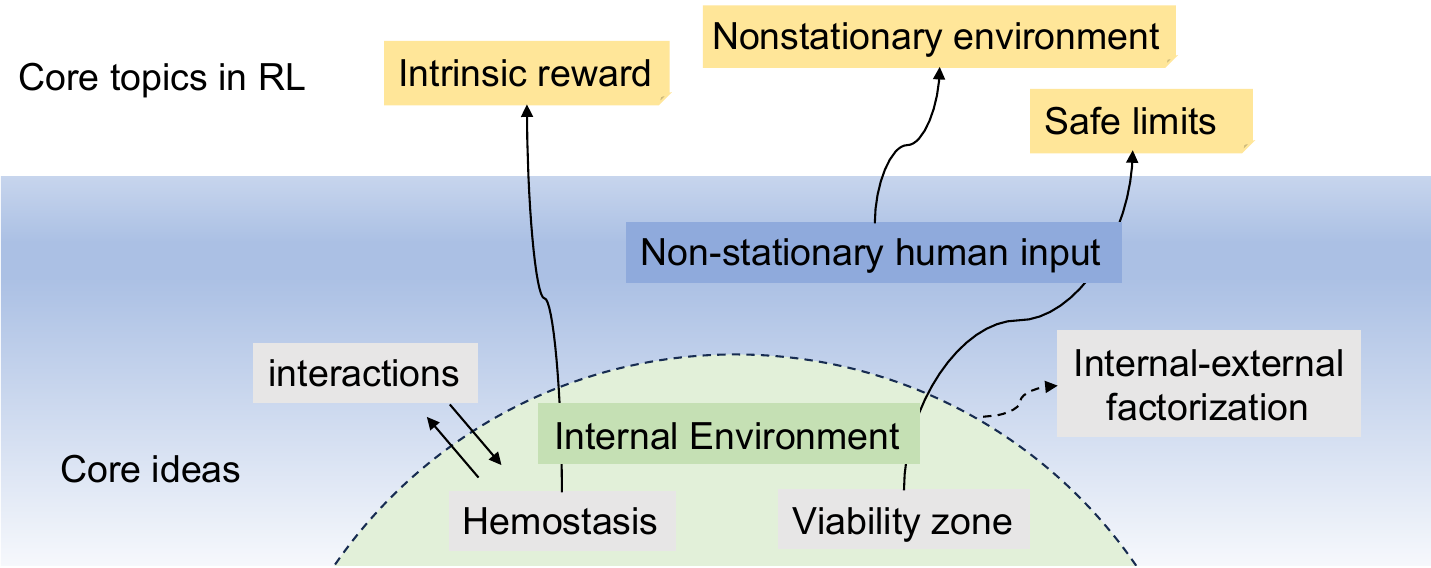}
    \caption{Overview of the interocepltive framework}
    \label{fig_2}
\end{figure}

With further decomposition, the knowledge required by robots for effective motion planning can be categorized into declarative knowledge and procedural knowledge. Declarative knowledge includes those facts (e.g., terrain features that were previously unavailable from the spatial grid map, traffic regulations with safety margins around obstacles, and path preferences provided by human supervisors) that inform future path-planning decisions. In our approach, this knowledge is acquired through human feedback that is prompted by an intuitive Graphical User Interface (GUI). The GUI facilitates the formalization of human knowledge while providing contextual dependency. This feedback is then encoded using natural language processing (NLP) techniques, including tokenization, tagging, and logical reasoning, and finally constructed as a knowledge graph (scaffolding). This design aligns with Lev Vygotsky’s theory of the Zone of Proximal Development, which highlights bridging the gap gradually between what the learner (robotic agent) can do independently and what it can do with guidance from a more knowledgeable other \cite{eun2019zpd}. Procedural knowledge pertains to how the mobile robot accurately replicates velocity profiles. Unlike dominant methods such as reinforcement learning (RL) and imitation learning—which start from a “blank slate”, rely on random exploration in large state spaces, or require explicitly designed rewards (a challenging task)—our approach formalizes learning as a structured process of schema representation and modification. This process is regulated by goal-oriented behavior, avoiding the habituation tendencies of RL that often fail to adapt quickly to reward contingencies and thus sacrifice flexibility \cite{khetarpal2022continual}. Our approach conceptualizes learning as a sequential production of components comprising a human-corrected velocity repertoire, implemented using an encoder-decoder architecture based on few-shot contrastive learning and Gated Recurrent Unit (GRU) to capture temporal dependencies. These designs lay the groundwork for factorizing the ontology of mobile robot motion planning and appeal to diverse optimization principles for routing decisions. An illustration of the robotic interoception concept, along with its relationship to modern topics in RL, is shown in Figure ~\ref{fig_2}.

\begin{figure}[!htb]
    \centering
    \includegraphics[width=0.48\textwidth]{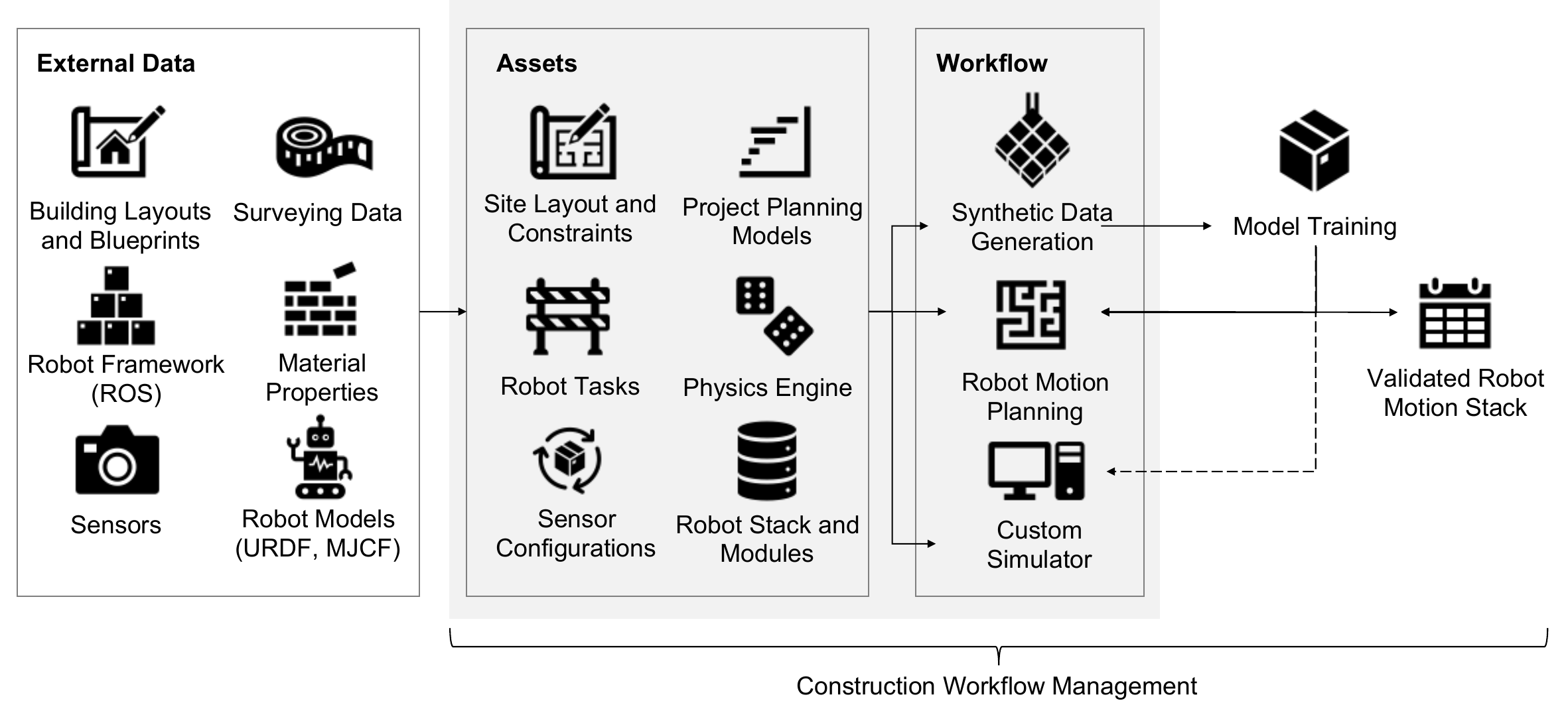}
    \caption{Simulation-enabled workflow for on-site robot motion planning}
    \label{fig_3}
\end{figure}

In practice, the workflow of the proposed approach is illustrated in Figure~\ref{fig_3}. While acknowledging the multifaceted nature of interoceptive robotics, this paper narrows its focus to a detailed methodology for motion planning in Section 2, followed by two illustrative examples—material delivery and handover placement—in Section 3.

\section{Methodology and Experimentation}
\label{sec:method and results}

\begin{figure*}[!htb]
    \centering
    \includegraphics[width=1\textwidth]{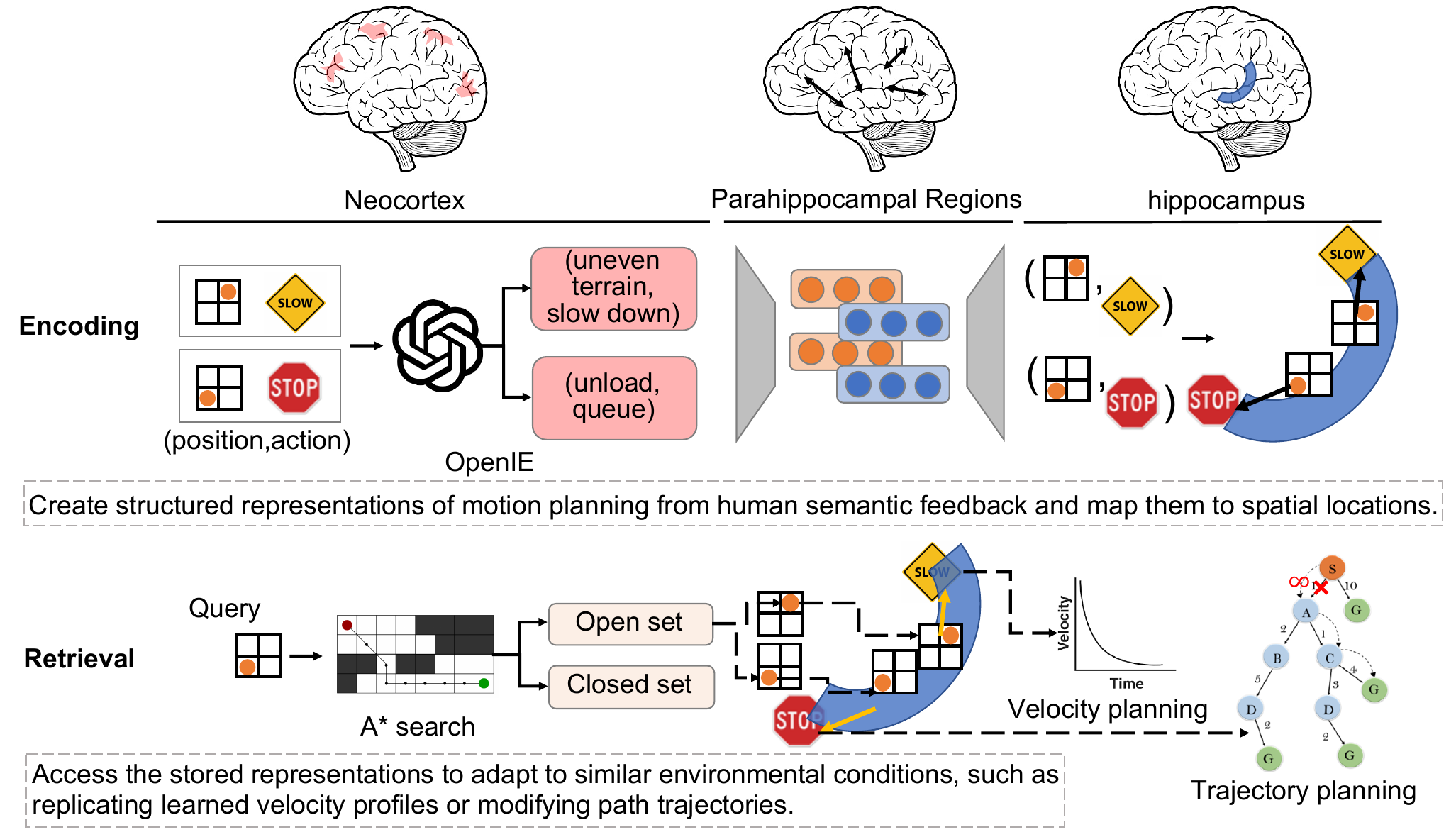}
    \caption{Overview of the proposed methodology}
    \label{fig_4}
\end{figure*}

The overarching pipeline of the designed methodology is shown in Figure ~\ref{fig_4}. The motion planning consists of two components: path planning and velocity planning. 

\subsection{Path Planning}

\begin{figure}[!htb]
    \centering
    \includegraphics[width=0.48\textwidth]{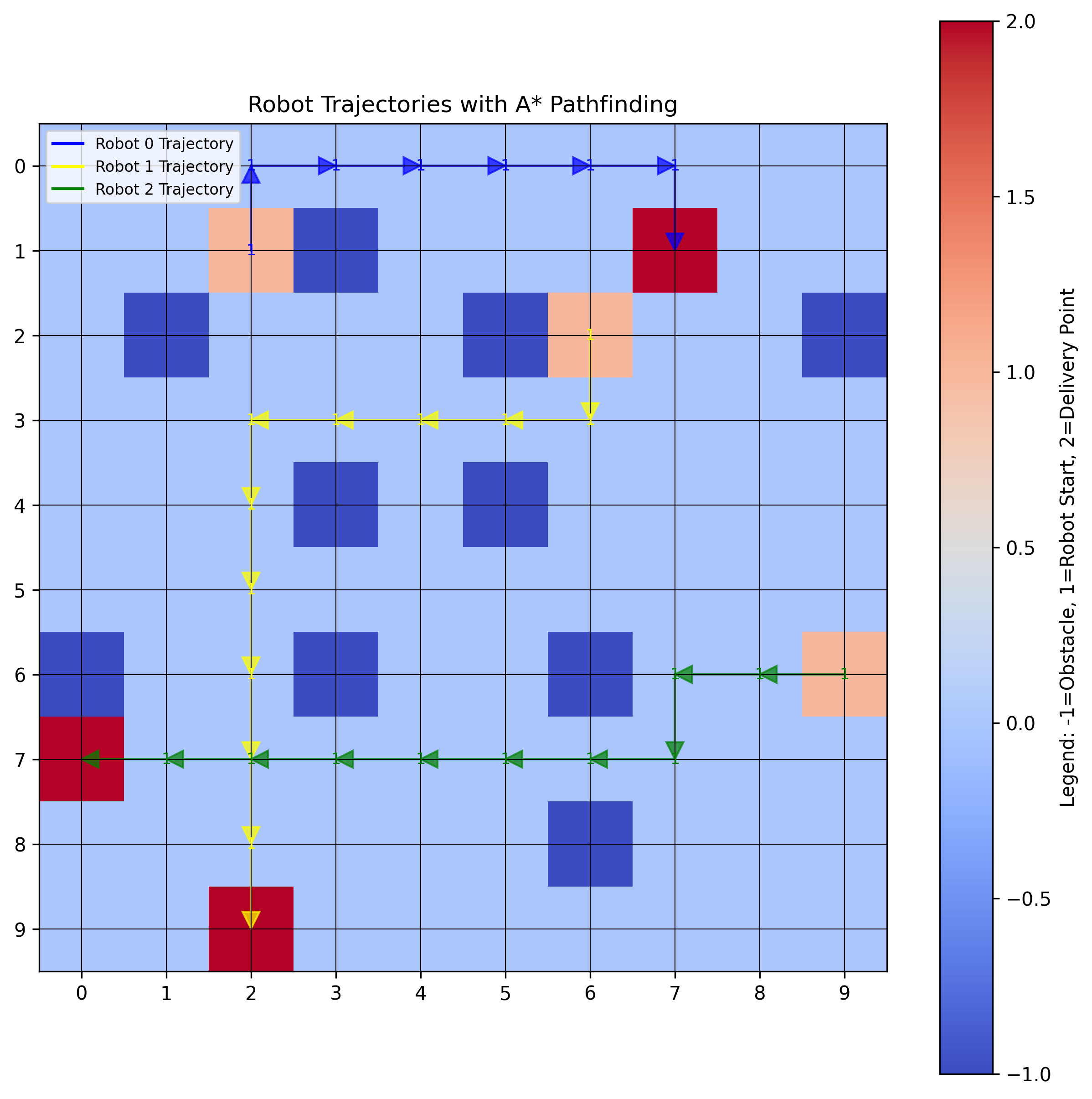}
    \caption{Trajectories with A* pathfinding in a grid}
    \label{fig_5}
\end{figure}

The path planning is defined to find a continuous path from a start location to a goal location throughout the subset of spatial configuration space consisting of traversable space. The space is created as a discretized approximation of connectivity using an occupancy grid, which can also be seen as a graph in which vertices are individual configurations and edges represent the transitions between pairs of configurations. Paths are found by searching the space using A* \cite{hart1968heuristic} (see Figure ~\ref{fig_5} for an abstracted formulation of the physical spatial environment and paths). 

\begin{figure}[!htb]
    \centering
    \includegraphics[width=0.48\textwidth]{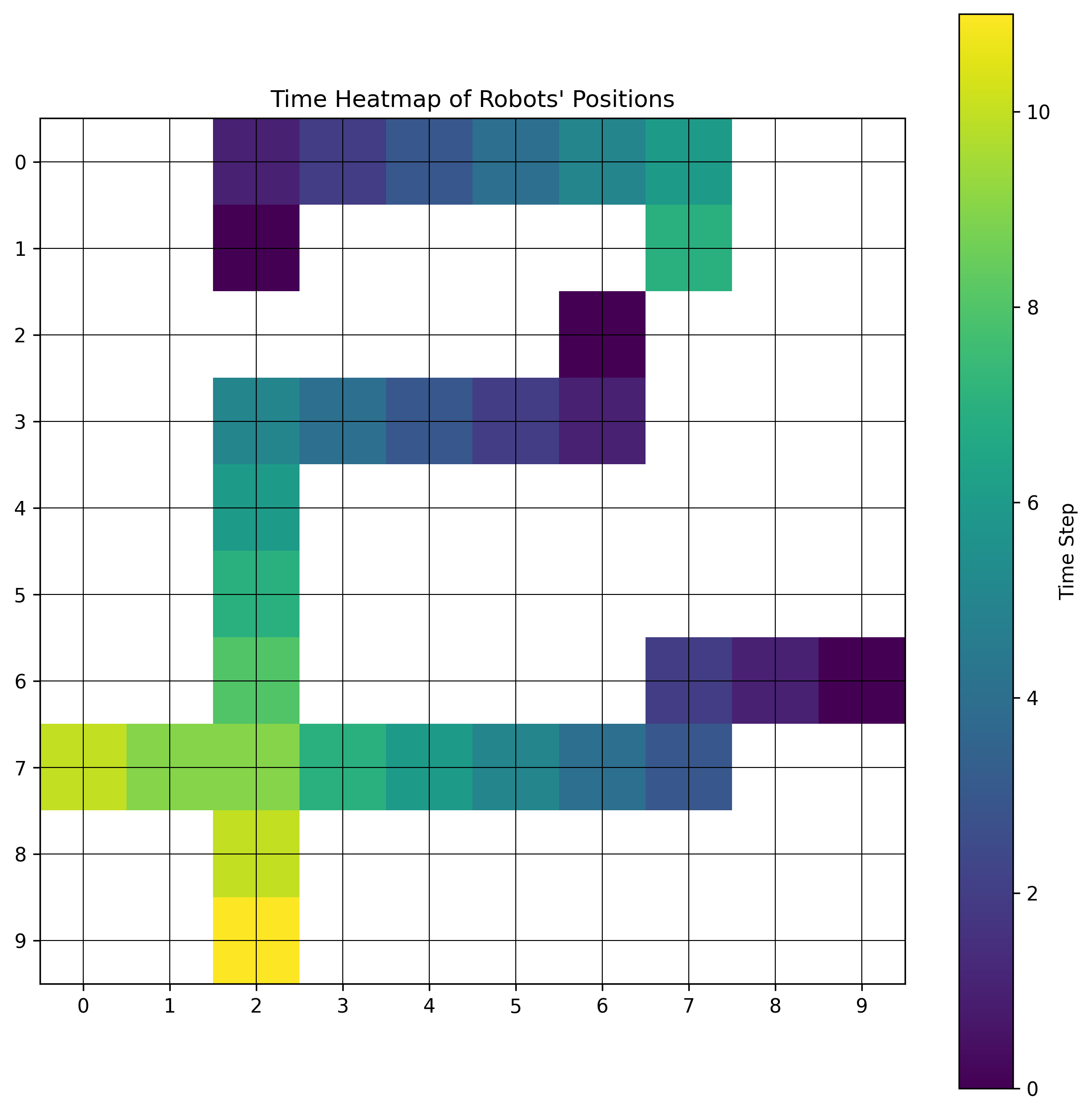}
    \caption{Heatmap of AMRs position tracking}
    \label{fig_6}
\end{figure}

\begin{figure}[!htb]
    \centering
    \includegraphics[width=0.48\textwidth]{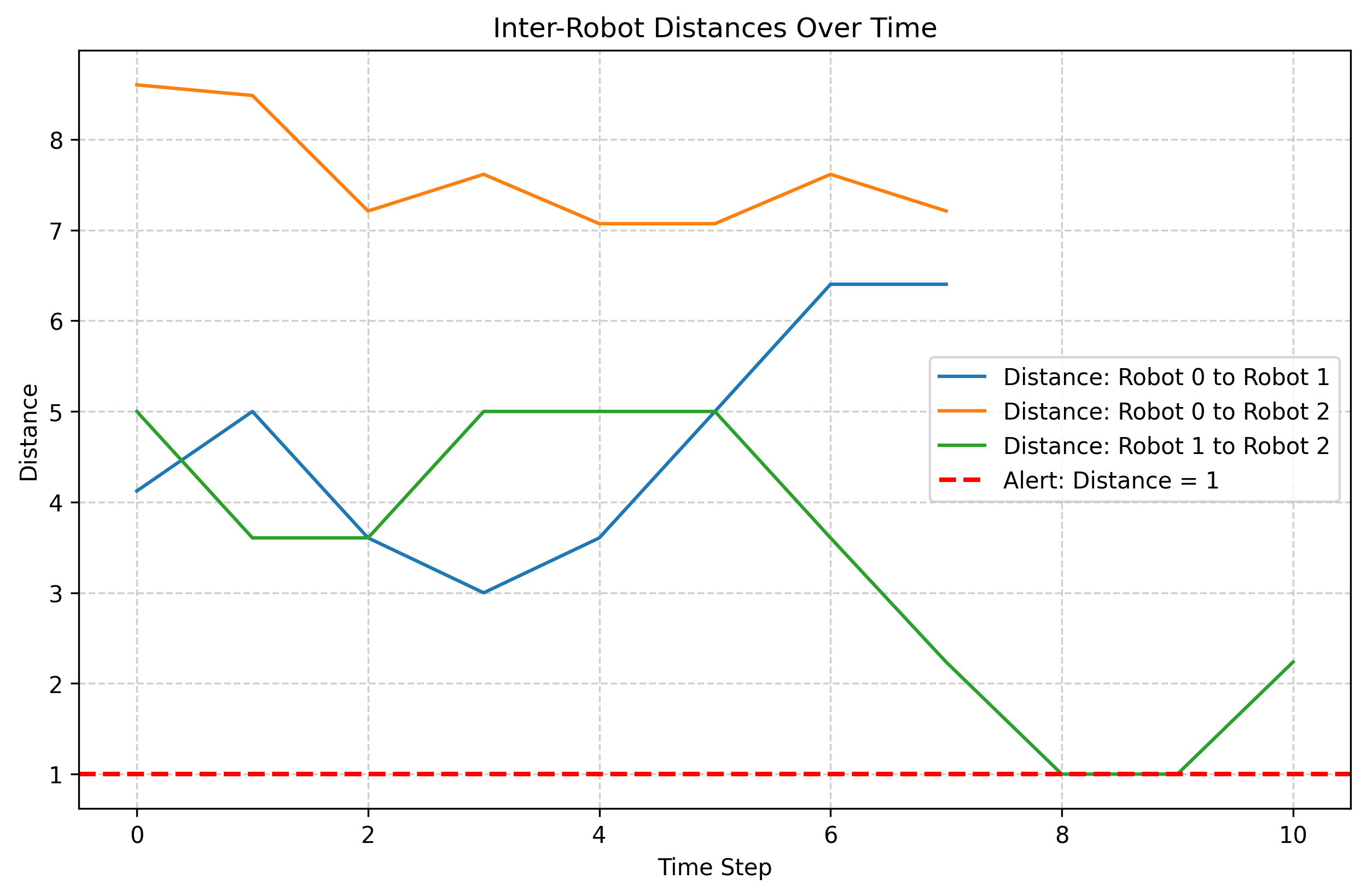}
    \caption{AMRs distance tracking and alerting}
    \label{fig_7}
\end{figure}

Given the designated paths, the tracking of the AMRs’ positions is available (see Figure ~\ref{fig_6}) with a hypothesized constant velocity. In practice, however, the human supervisors occasionally adjust the velocity to accommodate ground or floor conditions, task hierarchies, and dynamics with other entities. This research proposes a shared control framework with an average pooling of velocities from human agency (scaled as increments) and robotic autonomy (scaled in absolute terms). Using the integral of velocity curves (often approximated with sample-based approaches, e.g., Monte Carlo) at a pre-defined sampling rate along the designated path, the distances of the AMRs can be tracked in real time, with proximity thresholds set to trigger automated alerts (see Figure ~\ref{fig_7}).

\begin{figure}[!htb]
    \centering
    \includegraphics[width=0.48\textwidth]{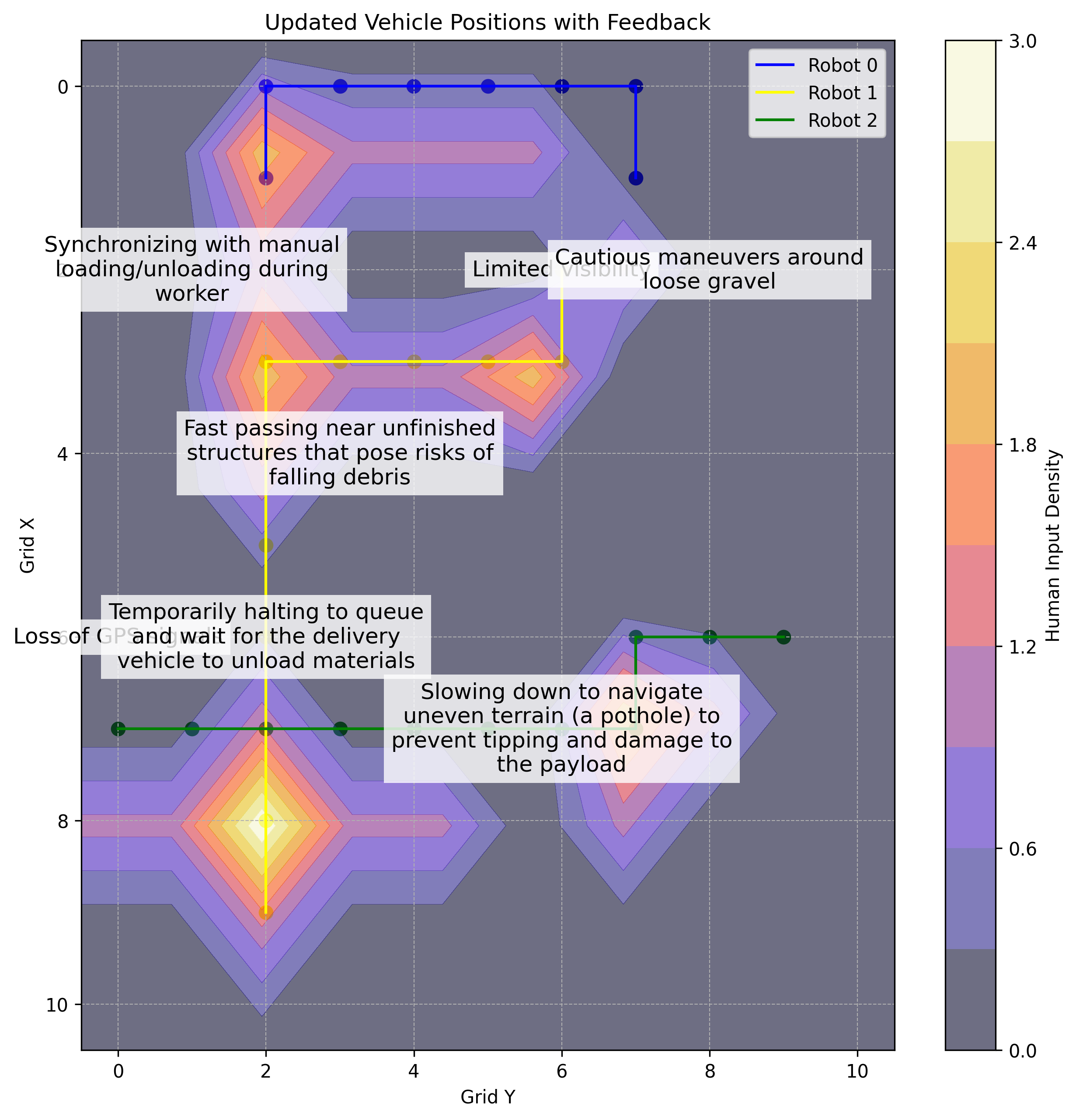}
    \caption{GUI for human semantic inputs}
    \label{fig_8}
\end{figure}

For closed-loop feedback, the factorized cognitive dissonance of mobile robot reasoning and action (as parameterized by the intensity of human inputs) is demonstrated using a contour plot, which provides an intuitive representation for retrospective documentation of rationale for human interventions (see Figure ~\ref{fig_8}). These texts are processed with NLP open information extraction (OpenIE) and archived focusing particularly on two attributes—terrain features and task sequences—as declarative knowledge mapped to the space vertices. Additional attributes are preserved as episodic knowledge, which is beyond the scope of this article. The source code for this simulation is available at \url{https://github.com/XiaoshanZhou624/A-star-with-GUI-for-human-inputs}.

\begin{figure}[!htb]
    \centering
    \includegraphics[width=0.48\textwidth]{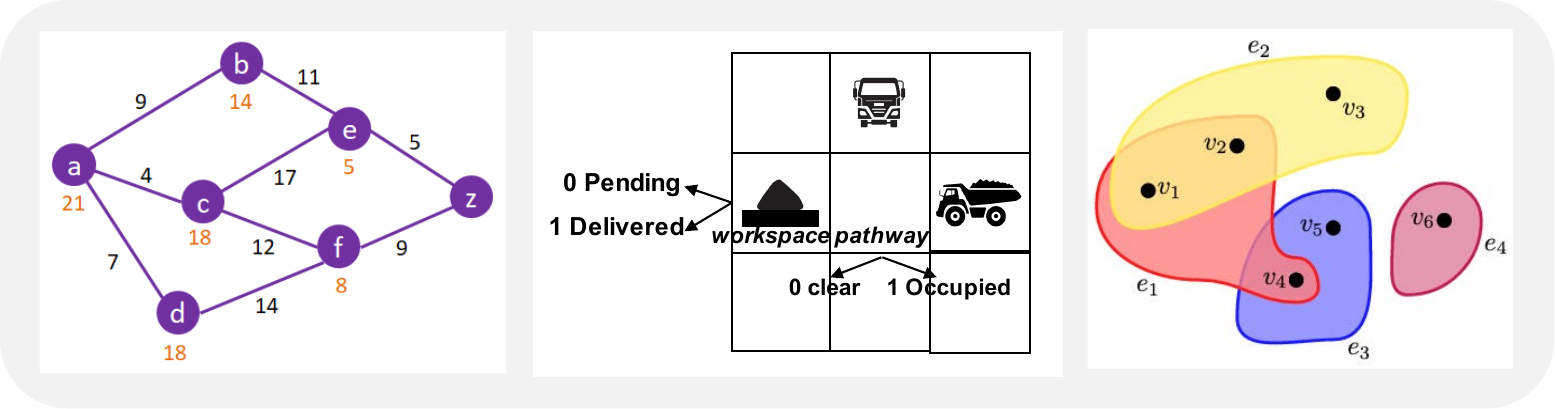}
    \caption{Hypergraph attribute visualization}
    \label{fig_9}
\end{figure}

\begin{figure*}[!htb]
    \centering
    \includegraphics[width=0.96\textwidth]{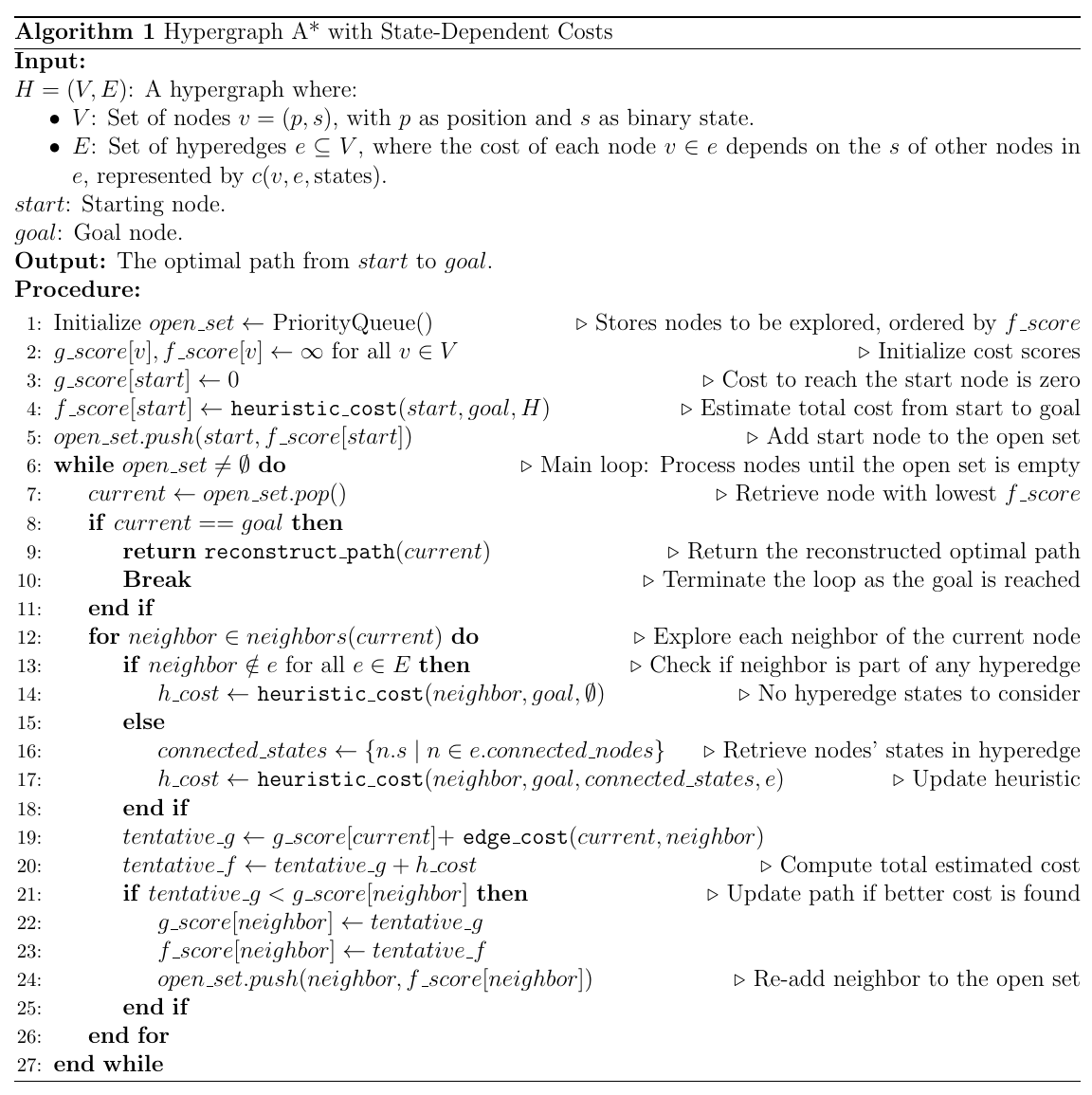}
    \caption{Pseudocode for hypergraph A* with state-dependent costs}
    \label{fig_10}
\end{figure*}

Task sequence attributes partially overlap with positional information from the spatial space but require an additional task space to define the availability state of the objects at the target workspace where robots offload or load and the occupancy state of critical pathways shared by robots approaching the workspace. To address this, we use hypergraphs to model this new configuration space with additional complexity. A hypergraph is an extension of a graph where hyperedges (or hyperarcs) are not restricted to connecting only two vertices \cite{berge1984hypergraphs}. This allows us to modify the heuristic costs in A*, automatically influencing path planning by determining which nodes are expanded as leaf nodes and added to the closed set. The task space states define the preconditions for admissible moves to the explored nodes. A visual depiction of the hypergraph is shown in Figure ~\ref{fig_9}, and the pseudocode for the modified A* algorithm is provided in Figure ~\ref{fig_10}.

\subsection{Velocity Planning}

\begin{figure}[!htb]
    \centering
    \includegraphics[width=0.48\textwidth]{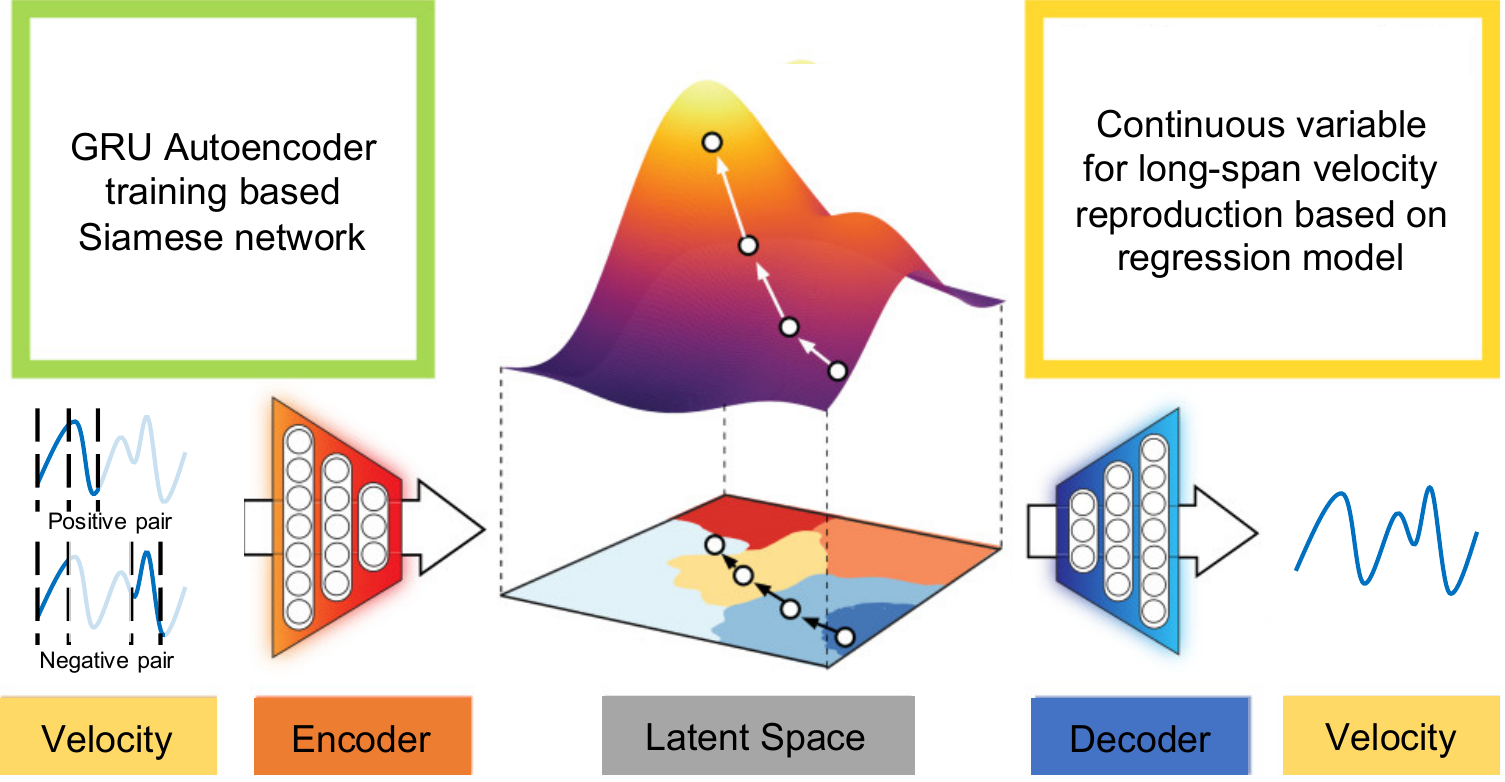}
    \caption{Technical overview of velocity encoding and retrieval}
    \label{fig_11}
\end{figure}

\begin{figure}[!htb]
    \centering
    \includegraphics[width=0.48\textwidth]{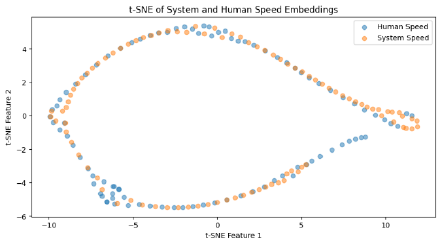}
    \caption{t-SNE of velocity embeddings}
    \label{fig_12}
\end{figure}

The spatial configuration is thus expanded, enabling the terrain attributes to trigger the replay of human-refined velocity profiles. This is achieved using an encoder-decoder network architecture, as shown in Figure ~\ref{fig_11}. Given only one trial of data, we designed a sliding time window to segment the velocity profiles into smaller segments. Leveraging the similarity contrast principles of few-shot learning, we construct positive pairs from consecutive windows and negative pairs from random windows. A Gaussian filter is applied to smooth the data within each window. Subsequently, these pairs are processed through a Siamese Network \cite{bromley1993signature} with GRU modules to extract embeddings for each window, using the Euclidean distance as the contrastive loss for network training. The embeddings are visualized in Figure ~\ref{fig_12}, with dimensionality reduced using t-Distributed Stochastic Neighbor Embedding (t-SNE). These embeddings are stored as compressed vectors of the original velocity profiles to improve information processing efficiency.

\begin{figure}[!htb]
    \centering
    \includegraphics[width=0.48\textwidth]{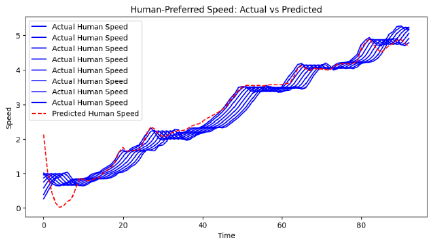}
    \caption{Illustration of velocity replay}
    \label{fig_13}
\end{figure}

During retrieval, a regression model is designed and trained using Mean Squared Error as the loss function to map the embeddings back to velocity values (see Figure ~\ref{fig_13} for a demonstration). These values can then be directly broadcasted and subscribed to by robotic servo motor nodes in the Robot Operating System (ROS) environment. Detailed algorithm designs are available in the open-source code repository at \url{https://github.com/XiaoshanZhou624/velocity-replay}.

\section{Demonstration in Simulation}
\label{third:demostration}

This section provides two examples of multi-robot coordination in motion planning and adaption to workpiece properties. The scenes and simulations were set up in Nvidia Omniverse Issac Sim (version 4.1.0).

\begin{figure*}[!htb]
    \centering
    \includegraphics[width=1\textwidth]{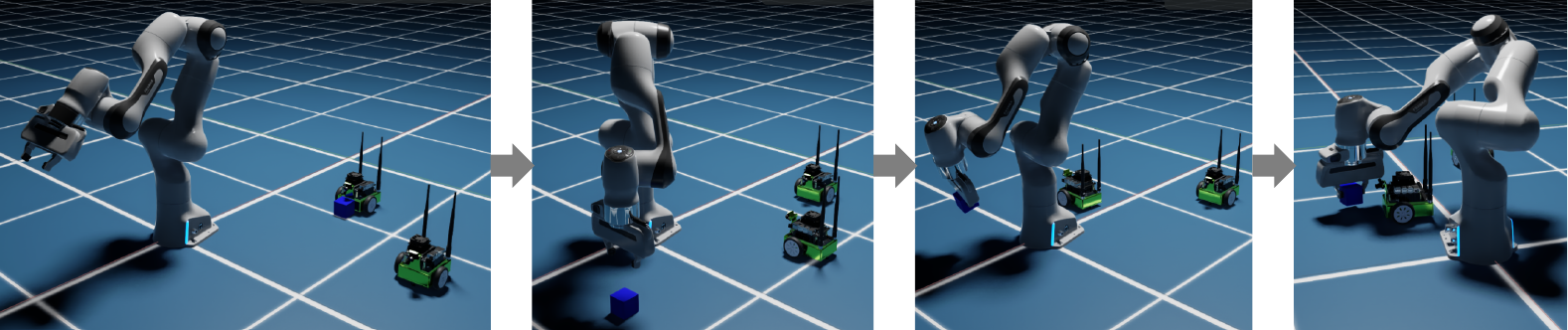}
    \caption{Robot motion planning for material delivery and handover coordination}
    \label{fig_14}
\end{figure*}

As shown in Figure ~\ref{fig_14}, two Jetbots were added to the simulation, acting as the preceding and following robots approaching the workspace. After the first Jetbot delivers material to the workspace, it must reverse to clear the pathway. Simultaneously, the second Jetbot must wait and avoid entering the workspace prematurely to prevent blocking the pathway. To represent the material, a cube was added to the scene as an abstracted object. A Franka Panda manipulator was positioned in the workspace to pick up the object and place it onto the second Jetbot that is responsible for transporting the material to another location. The physical properties were configured to enable the first Jetbot to push the material toward the Franka Panda manipulator for handover. The hypergraph vertices’ preconditions were incorporated into the program logic by parameterizing tasks or programming states to transition between subtasks, such as the Franka’s handover or allowing the Jetbot to resume movement instead of being blocked by non-steppable areas influenced by A* heuristic costs. A video of this simulation is available at \url{https://youtu.be/RwKJ0zggWfE}, and the Python scripts used to set up the scene, controllers, and task logic are available from the corresponding author upon reasonable request. In real-world applications, precise localization is often achieved using land-marked AprilTags for accurate positioning of robots and manipulators. However, as shown near the end of the video, when the goal location of the transport robot overlaps with the placement location of the manipulator’s end-effector, collisions can occur. This is particularly problematic with large automated machinery, such as forklifts, or when dropping large-scale material panels that require significant space for maneuvering, reorientation, or flipping before placement. These challenges highlight the importance of precise velocity control and synchronization of robotic motion to minimize downtime and ensure seamless operation.

In the simulation, the mobile robot’s controller was implemented using a differential drive mechanism, with robots constructed from physically accurate articulated joints. The robot’s speed was first converted into individual wheel speeds, which were then sent as commands to the joint drivers for articulation. Motion planning is conducted in the digital twin environment, allowing the plan to be tested and refined before execution in the real world. During simulation, human experts are enabled to interact with the system using peripheral devices such as gamepad controllers or keyboards. These human inputs are directly incorporated into the controller inputs, influencing the robot’s motion velocity in real-time. A data logging system was configured to record data including states (e.g., joint positions and orientations) and actions (e.g., linear and angular velocities), typically stored in JSON files. This logged data supports the learning framework presented in Section 2. When using keyboard controls, input scaling values were fine-tuned to map keyboard inputs to linear and angular velocities appropriate for the robot’s size and within a reasonable magnitude range.

\begin{figure*}[!htb]
    \centering
    \includegraphics[width= 1\textwidth]{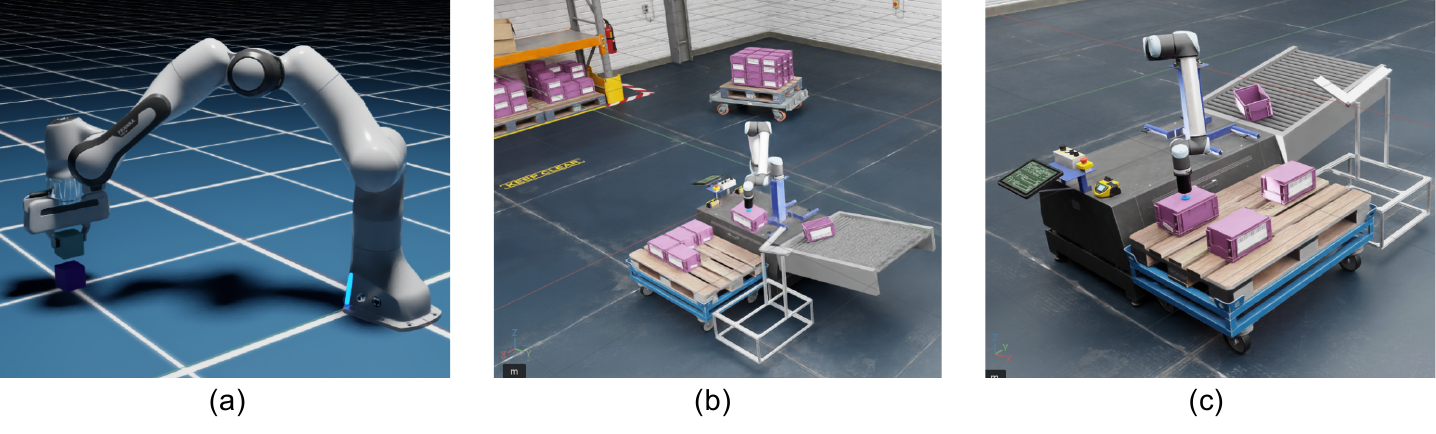}
    \caption{UR10 bin stacking and motion planning with weight-balanced placement}
    \label{fig_15}
\end{figure*}

As demonstrated, handovers often require precise calculations of space alignment and timing. Although block stacking tasks have been widely explored in robotic manipulation, real-world applications present greater complexity. Unlike uniform blocks with consistent shapes, sizes, and weights that can be easily aligned in horizontal or vertical stacks (see Figure ~\ref{fig_15}(a)), construction materials exhibit considerable variability. For instance, pieces of lumber often differ in dimensions, while palletized materials may contain different items, such as tiles and shingles, each varying in weight. Moreover, some materials may need repositioning during unloading to facilitate subsequent tasks. For example, glass panels may require flipping to expose the correct side for façade installation. Proper planning must account for balanced weight distribution during transport to mitigate the risk of tipping caused by uneven loads affecting inertia and centrifugal forces, particularly for wheeled robots (see Figure ~\ref{fig_15}(b)). Moreover, the coordination of multi-robot systems is critical for optimizing task scheduling, spatial allocation, and queue management throughout the onsite supply chain. The demonstrated system shown in Figure ~\ref{fig_15}(c), was adapted from a UR 10 Bin Stacking application containing a conveyor belt, a pallet where the bins should be stacked, and a UR 10 robot with a suction gripper. Bins arriving right-side up were flipped at a designated station to ensure proper stacking upside down. The placement alternated between sides of the pallet to balance weights. A supplementary video demonstrating the robot's motion planning is available at \url{https://youtu.be/yHbuoG7d05A}. Future work will further integrate sensors for weight tracking to enhance the motion planning of the manipulator, maintaining a balanced load while minimizing energy consumption and time during operation. In addition, sim-to-real discrepancies will be studied and minimized by analyzing the correlation between stiffness and damping parameters in the simulation setup (after importing robot models from URDF formats) and the variability of loads and terrain features that influence friction in the real-world. 

\section{Conclusion}
\label{fourth:conclusion}
We introduced the interoceptive robot framework, designed to enhance the autonomy and adaptability of robotic agents by incorporating an internal belief system and conflict-tracking mechanism within a new shared control architecture. The proposed framework enables the robotic agent to monitor its internal cognitive dissonance—arising from discrepancies between its self-belief in the decision space and the final actions convergent with human input. This interoception supports the agent’s ability to scaffold its learning by reflecting on contextualized rationale derived from human semantic feedback and thus adaptively recalibrate its motion planning to accommodate future similar environmental conditions. Importantly, this framework offers a fresh perspective on advancing robotized construction automation from rigid, pre-programmed behaviors to greater autonomy and adaptive intelligence. By learning from human counterparts in co-existent collaborative task hierarchies and under human supervision, it supports more heterogeneous and complex tasks within the modular construction vision, contributing to decarbonized and safer construction initiatives.

\section{Acknowledgement}
\label{fifth:acknowledgement}
The authors would like to acknowledge the financial support received for this research from the US National Science Foundation (NSF) Grant SCC-IRG 2124857. Any opinions in this paper are those of the authors and do not necessarily represent those of the NSF. 

\bibliography{ISARC}

\end{document}